\documentclass{easychair}

\usepackage{doc}
\usepackage{times}
\usepackage{latexsym}
\usepackage{graphicx}
\usepackage{amsmath}
\usepackage[linesnumbered,ruled,vlined]{algorithm2e}
\usepackage{multirow}
\graphicspath{ {/} }

\title{LexiPers: An ontology based sentiment lexicon for Persian}

\author{
Behnam Sabeti\inst{1}
\and
Pedram Hosseini\inst{2}
\and
Gholamreza Ghassem-Sani\inst{3}
\and
Seyed Abolghasem Mirroshandel\inst{4}
}

\institute{
Sharif University of Technology,
Tehran, Iran\\
\email{sabeti@ce.sharif.ir}
\and
University of Guilan,
Rasht, Guilan, Iran\\
\email{pdr.hosseini@gmail.com}\\
\and
Sharif University of Technology,
Tehran, Iran\\
\email{sani@sharif.edu}\\
\and
University of Guilan,
Rasht, Guilan, Iran\\
\email{mirroshandel@guilan.ac.ir}\\
}

\authorrunning{Sabeti, Hosseini, Ghassem-Sani and Mirroshandel}

\titlerunning{An ontology based sentiment lexicon for Persian}

\begin{document}

\maketitle

\begin{abstract}

Sentiment analysis refers to the use of natural language processing to identify and extract subjective information from textual resources. One approach for sentiment extraction is using a sentiment lexicon. A sentiment lexicon is a set of words associated with the sentiment orientation that they express. In this paper, we describe the process of generating a general purpose sentiment lexicon for Persian. A new graph-based method is introduced for seed selection and expansion based on an ontology. Sentiment lexicon generation is then mapped to a document classification problem. We used the K-nearest neighbors and nearest centroid methods for classification. These classifiers have been evaluated based on a set of hand labeled synsets. The final sentiment lexicon has been generated by the best classifier. The results show an acceptable performance in terms of accuracy and F-measure in the generated sentiment lexicon.

\end{abstract}



%
%


\section{Introduction}

Sentiment analysis (also known as opinion mining) is a field of natural language processing, which attempts to identify and extract subjective information from textual resources. Generally speaking, sentiment analysis aims at extracting topics that are discussed in a review or comment and then determining the attitude of the writer towards the extracted topic. Due to its efficiency and rapid growth in textual resources, sentiment analysis has a wide range of applications including customer review monitoring, political affiliation extraction, market movement prediction, etc. In other words, applications of sentiment analysis are becoming more popular every day. In order to extract subjective information, most researches have focused on automatically determining the polarity (e.g. positive or negative) of the terms in the given document. Our work, however, is dedicated to building a sentiment lexicon for Persian, a language spoken by approximately 110 million people worldwide. A sentiment lexicon associates each word with a sentiment orientation.

There are two types of sentiment lexicons:
\begin{itemize}
\item A general sentiment lexicon: each word is associated with a sentiment polarity; for example, ''good'' is a positive word.
\item A domain specific sentiment lexicon: each word is associated with a sentiment polarity w.r.t a specific domain. For example ''unpredictable'' is a positive adjective if used for a movie, but it expresses a negative orientation towards a politician.
\end{itemize}

In this research, we have focused on building a general sentiment lexicon for Persian.

There are three approaches for sentiment lexicon generation:
\begin{itemize}
\item Manual approach: manually labeling all words w.r.t their orientations.
\item Dictionary based approach: using a dictionary (e.g. WordNet) and its relations between words to generate a sentiment lexicon.
\item Corpus based approach: employing words co-occurrences for assigning sentiment polarity to each word. For example if a word occurs in a pattern like ''good and X'' then ''X'' is more likely to be a positive sentiment word.
\end{itemize}

FarsNet, a publicly available ontology in Persian containing more than 20,000 synsets \cite{shamsfard2010semi}, has been used as the basis of our developed sentiment lexicon. In this work, lexicon development has been done as a document classification task. More specifically, given each synset, the glossary associated with it, is classified as a positive, neutral, or negative document \cite{esuli2005determining}.

Our approach is in fact a combination of all three mentioned approaches for lexicon generation. At first, graph analysis is employed for selecting a set of words. The selected set of words (also known as seed list) is then manually labeled and expanded using FarsNet and a PMI-based method. The expanded list is finally used as a training set for the document classification task.

In order to evaluate our proposed method, a set of synsets has been manually labeled. Different classifiers have been trained on several training sets and evaluated with respect to the test set. The best classifier and training data is then used for lexicon generation.

The rest of this paper is organized as follows. In section \ref{sec:relatedWork}, we mention some of the most important researches related to the development of polarity lexicons for English. We also discuss the majority of works that have been done for generating sentiment lexical resources in Persian. Then, the process of developing our lexicon, named LexiPers, discussed in detail in section \ref{sec:development}, in addition to introducing our proposed method. A detailed account of different steps in the development of LexiPers from creation of the initial seed to learning process, and document classification will be presented in section \ref{sec:experiment}. Section \ref{sec:conclusion} concludes the paper.

\section{Related Work}
\label{sec:relatedWork}

One of the earliest polarity lexicons that have been built is General Inquirer (GI) \cite{stone1966general}. GI lexicon is part of the GI system, a content analysis program that exploits terms manually classified on various categories. It includes a list of terms from Harvard and Lasswell. For each term in this list, there is a positive, negative, or no label assigned from the subjectivity point of view.

Multi-perspective Question Answering (MPQA) is another known subjectivity lexicons with prior polarities \cite{wilson2005recognizing}. It is part of OpinionFinder \cite{wilson2005opinionfinder} and contains over 8,000 subjectivity clues. Each word has a type: ''strongsubj'' for those words that are subjective in most contexts, and ''weaksubj'' for those that may only have certain subjective usages. In addition to type, each word has a prior polarity of positive, negative, or neutral.

Another well-known and recent sentiment lexicons for English is SentiWordNet (SWN) \cite{esuli2006sentiwordnet}. In SentiWordNet, three scores are assigned to each of WordNet synsets showing how positive, negative and objective these synsets are. These scores are derived from combining the results of eight ternary classifiers. The major idea behind constructing SentiWordNet is the classification of WordNet synsets using semi-supervised classification of synset glosses. An important point that should be considered regarding the scores assigned to each synset is that these scores do not show the strength of the polarity. They only show that how much a synset is going to be positive, negative or objective. As a result, the sum of these three scores for every synset is equal to 1.0.

Another version of SWN includes a random-walk step for refining the scores besides the semi-supervised learning step of the earlier version \cite{baccianella2010sentiwordnet}.

SentiFull is another work that uses an automatic method for generating and scoring a sentiment lexicon \cite{neviarouskaya2011sentiful}. In this work some relations including synonymy, antonymy and hyponymy have been used for lexicon expansion in addition to derivation and compounding with known lexical units. The accuracy of scores has also been evaluated using a randomly selected list of terms from the generated lexicon annotated by human annotators.

In another work, a sentiment lexicon has been constructed automatically with domain specific polarity for terms \cite{lu2011automatic}. In addition, in a specific domain, differences between polarities of a word based on its aspects in the context was also taken into the consideration. In other words, depending on the aspects of each word in a context, different polarities may be assigned to that word.

Like many other languages, working on sentiment analysis in Persian is growing fast. However, there are rather limited works dedicated to building sentiment analysis resources and datasets for Persian. We cover the work related to developing sentiment \textit{lexicons} for Persian in the following paragraph.

Persian Clues is a sentiment lexical resource that has been developed for Persian \cite{shams2012non}. Two main steps are taken for building Persian Clues. An initial set of clues is collected using an existing English lexical resource in the first step \cite{wiebe2005creating}. Then, in the second step, errors in the initial seed are removed and final clues are created. In another work, a sentiment lexicon for Persian is created using the SentiWordNet and the Persian wordnet named FarsNet \cite{alimardaniopinion}. The idea behind building the lexicon is mapping each SentiWordNet entry to its equivalent in FarsNet (if available) using WordNet. In another study, a Persian sentiment lexicon has been generated using an existing English lexical resource, Subjective Clues \cite{golpar2015feature} \cite{wiebe2000learning}. The method for building the lexicon is based on translation and is identical to previously mentioned methods for developing Persian sentiment lexicon.

\section{Process of Developing LexiPers}
\label{sec:development}

\begin{figure}[tb]
\centering
\includegraphics[width=0.5\textwidth]{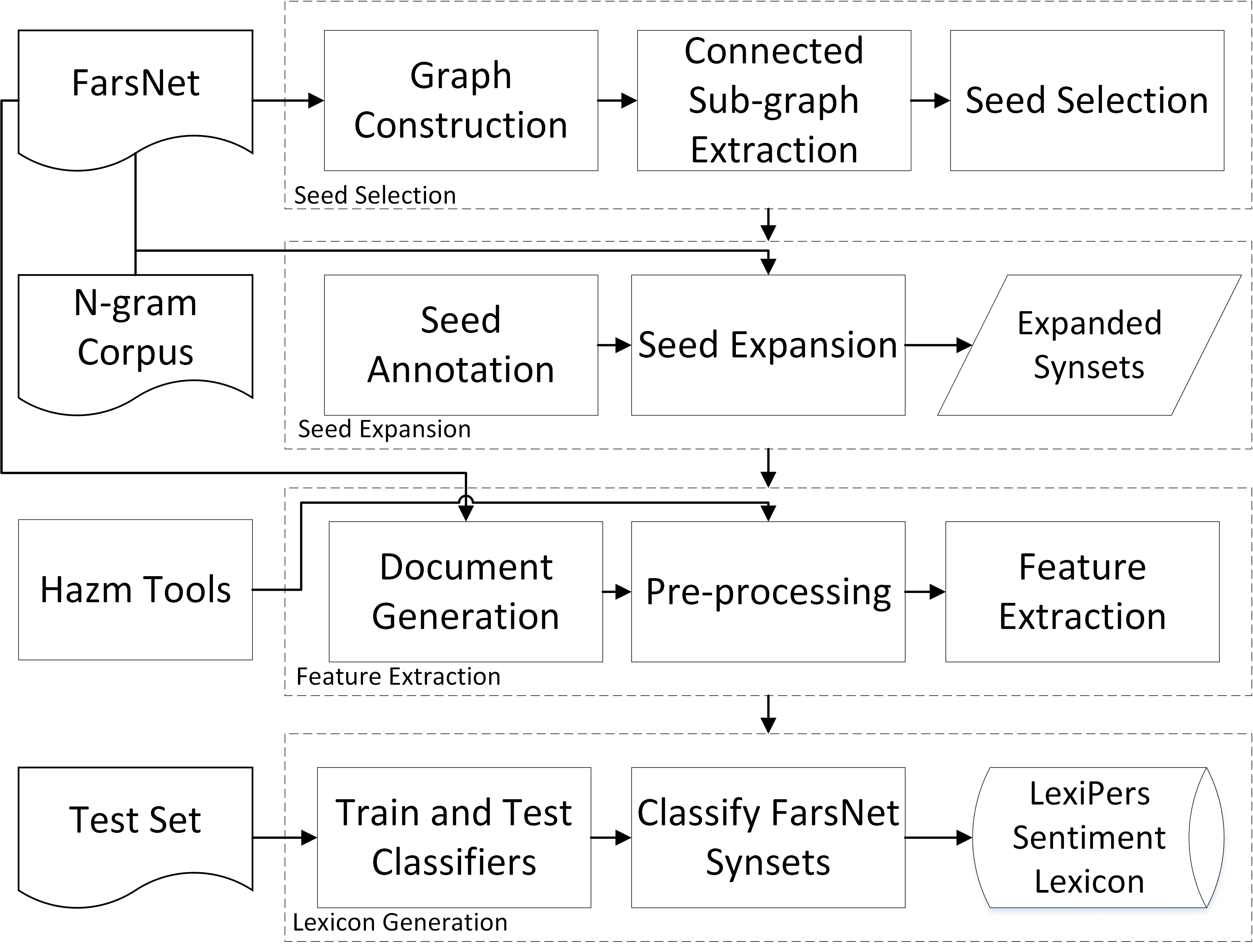}
\caption{Proposed Method Architecture}
\label{fig:Architecture}
\end{figure}

An overview of our proposed method is shown in Figure \ref{fig:Architecture}. In this part we explain the entire process in detail.

As it was mentioned before, we have generated a sentiment lexicon based on the FarsNet ontology. Each synset in this ontology is described with a glossary. Synsets are classified based on their glossary. Here, we introduce a new approach for seed selection and expansion. Document classification is then employed for lexicon generation.

\subsection {Creating an initial seed}
The first step toward training the required classifier is to select the seed synsets. These synonym sets will then be tagged and expanded to form the training set. In this work, as opposed to previous sentiment lexicons \cite{esuli2006sentiwordnet}, seed selection is performed as a separate process to ensure the best result after expansion. An ontology is used to build a graph, modeling the words and their relations in Persian. Graph analysis is then employed to select the seed list in a way that would guarantee the generation of the best training set after expansion.

We are in general interested in a seed list that is as small as possible and is converted to a set of nodes that is as large as possible after the expansion phase. This way we annotate a small set of nodes and expansion phase will provide us with a large set of annotated nodes. Using the following theorem, we can find a minimum set of nodes that will be expanded to all of the nodes in the graph.

\textbf{\textit{Theorem:}} In order to select the initial seed list, convert the ontology to its corresponding graph. Extract all of the strongly connected sub-graphs from the generated graph. From each sub-graph add a single node to the initial seed list. Using this initial seed list, after expansion phase all of the nodes in the graph will be annotated.

\textbf{\textit{Proof:}} In the expansion phase, in each step all of the neighbors of the seed list will be annotated based on the relations in the ontology. Furthermore, if there is a path between the seed list and any nodes in the graph, that node will be annotated automatically. We select the seed list based on the extracted strongly connected sub-graphs. By definition in a sub-graph there is a path between any two nodes. So each node in the initial seed list guarantees that its corresponding sub-graph will be annotated after expansion since it has a path to each node in that sub-graph. Finally we conclude that after expansion phase every node in the seed list leads to annotating the nodes in its corresponding sub-graph and this way all of the nodes in the graph will be annotated.

\subsubsection {Graph construction}
In this section, we explain how we constructed a graph representing the FarsNet ontology. To this end, each synset in the ontology is mapped to a node in the graph. Since ontology relations are defined between two synsets, each relation can be mapped to an edge in the graph.

Although we use all the synsets to generate the nodes, only antonym relations are considered as edges in the graph. This is because we are interested in the sub-graphs with sentiment expressing nodes, and nodes participating in antonym relations are more likely to express sentiment orientations.

\subsubsection{Finding strongly connected sub-graphs}
In order to select the initial seed list, strongly connected sub-graphs should be extracted. In a strongly connected sub-graph, each pair of nodes are connected through at least one path. There are several algorithms for this purpose. We have employed the Kosaraju algorithm for strongly connected sub-graph extraction \cite{hopcroft1983data}.

\subsubsection{Seed selection and annotation}
The FarsNet graph has been divided into a set of strongly connected sub-graphs. A single node is then selected from each sub-graph and added to the initial seed list. This seed list is guaranteed to be the smallest set of nodes that after expansion leads to the largest expanded list.

The selected seed list has to be annotated by hand. For each node corresponding to a synset, its sentiment orientation should be specified. In this work, two trained annotators, both native Persian speakers with proper knowledge and understanding of sentiment analysis concepts have annotated the seed list. We have also measured the agreement among these annotators using Fleiss` kappa \cite{fleiss1971measuring} and the result shows 90\% agreement. It is worth pointing out that for any synset with disagreement between the annotators, a third person has annotated the synset and his judgement for the polarity has been chosen as the final label of the synset.

\subsection {Seed expansion}
\label{sec:seed-expansion}

\begin{figure}[tb]
\centering
\includegraphics[width=0.5\textwidth]{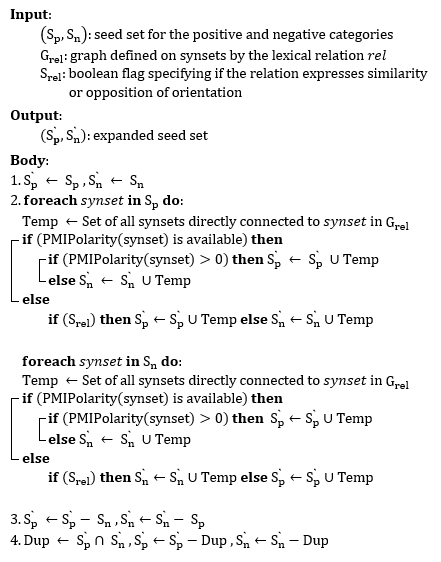}
\caption{The bootstrapping algorithm for seed expansion using PMI}
\label{fig:expansion}
\end{figure}

After creating and annotating an initial seed list, the list is expanded using a bootstrapping method in two different cases. In the first case, we used an algorithm identical to \textit{ExpandSimple} function that was introduced by Esuli and Sebastiani in 2005. In the second case, using Pointwise Mutual Information (PMI) \cite{church1990word}, we applied some minor changes to the \textit{ExpandSimple} function to see if we could improve the accuracy of polarity labels during seed expansion process. The Pseudo-code of our bootstrapping algorithm, \textit{ExpandSimpleByPMI}, is shown in Figure \ref{fig:expansion}. There are some differences between \textit{ExpandSimple} and our algorithm, \textit{ExpandSimpleByPMI}. First, our initial and expanded seeds are a set of positive and negative \textit{synsets} not \textit{terms}. Second, the polarity of a synset is assumed to be equal to the sum of the polarities of senses of the synset. Finally, the key difference between \textit{ExpandSimple} and \textit{ExpandSimpleByPMI} is in the way that these algorithms assign a polarity to the synsets. In the former function, it is assumed that Synonymy, Hypernymy, and Hyponymy relate terms with the same orientation, while Antonymy relate terms with opposite orientation. However, in the latter one, except for the first round of bootstrapping, we first calculate the PMI polarity of the new synset using the formula introduced by Turney and Littman for finding the orientation of a term \cite{turney2003measuring}. We have used a list of unigrams and bigrams from Parsijoo.ir, a Persian search engine. Also, we considered \textit{good} and \textit{bad} as our positive and negative seed sets, respectively. After calculating the PMI polarity, we assign the polarity of the synset based on this value. For PMI greater or less than zero, the polarity of the synset is +1 and -1, respectively. In those cases that PMI is not available, we use the assumption of the default algorithm. By availability of the PMI, we mean that we have all the required unigrams and bigrams so that we can calculate the sentiment orientation of a synset. We do not apply \textit{ExpandSimpleByPMI} for assigning polarities at the first round because the synsets in the first round of bootstrapping are directly connected to our hand-labeled synsets.

In the bootstrapping step, we considered a few parameters and ran the algorithm for a number of times changing the value of these parameters. After all, we came to three different cases for the bootstrapping process. These cases are shown in Table \ref{table:expansion}. When the value of \textit{All POS} is equal to \textit{Yes} it means that the synsets with part of speeches among nouns, verbs, adjectives or adverbs have been taken into the consideration in the bootstrapping process. Otherwise, only synsets with adjective part of speech have been used in the seed expansion.

\begin{table}
\centering
\small
\caption{Different states of seed expansion}
\label{table:expansion}
\begin{tabular}{|c|c|c|c|c|}
\hline
\multicolumn{2}{|l|}{} & \multicolumn{3}{c|}{Relation} \\ \hline
\multicolumn{1}{|l|}{Case} & All POS? & Antonym & Hyponym & Hypernym \\ \hline
1 & Yes & Yes & No & No \\ \hline
2 & No & Yes & Yes & Yes \\ \hline
3 & No & Yes & No & No \\ \hline
\end{tabular}
\end{table}

\subsection {Learning process}

In the first two steps a semi-supervised method was employed for annotating a set of FarsNet synsets. These annotated synsets can then be used as a training set for classifying the rest of the synsets.

As it was mentioned before, each synset consists of a set of relations and a glossary. The problem of classifying these synsets can be mapped into the problem of classifying theirs corresponding glossaries. Thus, document classification can be employed to classify these synsets. Finally, the trained model is used for classifying the rest of the synsets in FarsNet, generating the final sentiment lexicon.

\subsubsection {Document Generation}

In our approach, each synset in the FarsNet ontology had to be converted to a document. Each FarsNet synset consists of a set of senses, a glossary, a set of examples, and a set of relations.

Synset glossary is a brief description of each synset. In order to use all of the information in each synset, we concatenated the senses and examples with the glossary. Therefore the concatenation of the glossary, senses, and the examples formed the document for each synset.

\subsubsection{Pre-processing Documents}
After generating the documents, we performed some preprocessing steps before extracting features. In particular, the preprocessing phase consisted of three stages, including:

\begin{description}
\item[Tokenization:] In the second step, we performed two different types of tokenization including sentence tokenization and word tokenization.
\item[Stemming and Lemmatizing:] After tokenizing the document into its tokens, we lemmatized and stemmed each token of the document. Lemmatizing and stemming result in reducing the size of the features in our vector representation.
\item[Part-of-speech tagging:] Finally, we ran the process of part-of-speech (POS) tagging. Using POS, we removed those tokens that seemed not to play an important role in polarity detection and kept those tokens that were going to carry a specific polarity. For instance, adjectives and nouns usually are assumed to have a stronger polarity, compared to other part of speeches.
\end{description}

\subsubsection{Feature extraction}

In order to classify the generated documents, each document had to be converted to a feature vector. We used the term-frequency-inverse-document-frequency (also known as TF-IDF) as the features for each document.

TF-IDF is a frequency based method for feature extraction. Furthermore, it counts the number of appearances of each word in the document with the advantage of scaling the frequencies with respect to the frequencies in the entire set of documents. This way the words that occur in most of the documents are assigned a low weight whereas the words occurring in some specific documents are emphasized with high weights. This weighting scheme implements our intuition of the importance of each word in each document. Furthermore, we are interested in the words that only appear in some specific documents because these words can hint the class of each document, as opposed to the words like ''of'' or ''the'' that appear in every document and are useless for document classification in the bag of words models.

\subsubsection{Training classifiers}

So far we have generated a set of vectors and their corresponding labels (e.g. positive, neutral, or negative). These vector and label pairs are then used to train a classifier for synset classification.

We used k-nearest neighbors (also known as KNN) and nearest centroid (also known as Rocciho) for classification. The intuition behind selecting these classifiers will be explained later in the paper. After parameter estimation, these classifiers were trained using the training data generated from the expanded seed list. Each classifier was then evaluated with a hand labeled test set.

\subsection {Classifying synsets}

After training two classifiers for Farsnet synsets, The superior classifier was selected based on a hand labeled test set. The chosen classifier was then employed for Farsnet synset classification. Finally all the synsets and their corresponding sentiment polarity formed the LexiPers sentiment lexicon.

\section{Experiments and results}
\label{sec:experiment}

This section illustrates the results of our proposed method. First, seed selection and expansion results are presented, then classifiers are modeled based on the expanded seeds. Classification evaluation ends this section.

\subsection{Seed selection and expansion}

As it was mentioned in the proposed approach, the first step of our algorithm was to convert the FarsNet ontology to its corresponding graph. Each node in this graph corresponds to a synset in the ontology. Relations between the synsets are considered as the edges in the graph. We used antonym relations to construct our ontology graph.

The Kosaraju algorithm is then employed to extract connected sub-graphs. Table \ref{table:subgraph_size} illustrates the size of these extracted sub-graphs. A sub-graph size is the number of nodes contained in that sub-graph.

\begin{table}[ht]
\centering
\small
\caption{Details of the extracted strongly connected sub-graphs}
\label{table:subgraph_size}
\begin{tabular}{|c|c|c|c|c|}
\hline
sub-graph size (node) & 1 & 2-5 & 6-10 & 11-15 \\ \hline
number of sub-graphs & 18351 & 805 & 21 & 4 \\ \hline
\end{tabular}
\end{table}

The constructed FarsNet graph is composed of 18,351 single-node sub-graphs. Annotating these nodes will not add any other synsets in the expansion phase. On the other hand, the main purpose of this research has been generating a sentiment lexicon automatically, but annotating this number of synsets would not be feasible.

In this step, sub-graphs with more than 2 nodes are selected for annotation. Since in a strongly connected sub-graph, there is a path between each pair of nodes, therefore, one node is selected randomly from each sub-graph for annotation. The expansion phase will cause all the nodes in each sub-graph to be annotated.

As it was mentioned in the section \ref{sec:seed-expansion}, different combinations of FarsNet relations were used for the expansion phase, leading to 3 expansion cases. The annotated seed list is expanded with respect to each expansion case and each expansion algorithm. This procedure leads to six training datasets, which are then used in the document classification phase. Table \ref{table:train-data} shows the details of each training dataset.

\begin{table}[ht]
\centering
\small
\caption{Training Datasets details}
\label{table:train-data}
\begin{tabular}{|c|c|c|}
\hline
State & Expansion algorithm & Dataset \\ \hline
\multirow{2}{*}{1} & Default & Data-1 \\ \cline{2-3}
& PMI & Data-2 \\ \hline
\multirow{2}{*}{2} & Default & Data-3 \\ \cline{2-3}
& PMI & Data-4 \\ \hline
\multirow{2}{*}{3} & Default & Data-5 \\ \cline{2-3}
& PMI & Data-6 \\ \hline
\end{tabular}
\end{table}

\subsection{Document classification}

The expanded seeds can now be used as the training samples for classification. As it was mentioned earlier, each node in these seeds had to be firstly converted to a document. After this conversion, there would be a set of documents and their corresponding class labels (e.g. positive, neutral, or negative).

\subsubsection{Classifier selection}

In the process of selecting an appropriate classifier for document classification, we observed that the resulting training sets are highly unbalanced; furthermore training sets were mostly consisting of neutral documents. In order to overcome this problem, we decided to use sample based classifiers (also known as lazy classifiers).

Our experiments show that, due to the omission of the training phase, these classifiers can generate better approximations for new documents. Therefore K-nearest neighbors (also known as KNN) and nearest centroid (also known as Rocciho) were selected as our classifiers in this research.

\subsubsection{Classification results}

As mentioned in the seed expansion section, we expanded our seed list in different ways to explore the effects of each FarsNet relation on the final results.

Prior to the usage of these classifiers, parameters of each classifier should be tuned based on the training sets. KNN parameters are the number of neighbors (or k), and the distance measure used to compare samples. Whereas, Rocciho has only a similarity measure parameter, which needs to be set. 10 fold cross validation is employed for parameter estimation.

A total number of 2861 synsets have been annotated manually and used as the gold standard for evaluating the document classification. The same two annotators of the initial seed list annotated these synsets. The agreement among these annotators using Fleiss` kappa was about 76\%. After parameter estimation, each classifier was evaluated with our gold standard. Table \ref{table:classification-results} illustrates accuracy and f-measure of each classifier using each training set.

\begin{table*}[t]
\centering
\small
\caption{Classification Results}
\label{table:classification-results}
\begin{tabular}{|c|c|c|c|c|}
\hline
\multirow{2}{*}{Training set} & \multicolumn{2}{|c|}{KNN} & \multicolumn{2}{|c|}{Rocciho} \\ \cline{2-5}
& \multicolumn{1}{|c|}{Accuracy} & \multicolumn{1}{|c|}{F-measure} & \multicolumn{1}{|c|}{Accuracy} & \multicolumn{1}{|c|}{F-measure} \\ \hline
Data-1 & 0.76226 & 0.58169 & 0.78432 & 0.60629 \\ \hline
Data-2 & 0.7564 & 0.57385 & 0.16968 & 0.35815 \\ \hline
Data-3 & 0.80826 & 0.66421 & 0.79465 & \textbf{0.663} \\ \hline
Data-4 & 0.80357 & 0.65432 & 0.1847 & 0.35268 \\ \hline
Data-5 & 0.81061 & \textbf{0.66708} & 0.78245 & 0.60467 \\ \hline
Data-6 & 0.80544 & 0.65638 & 0.79277 & 0.66073 \\ \hline
\end{tabular}
\end{table*}

Both classifiers reached almost the same f-measure at their best performance. Using the datasets with the default expansion algorithm, Rocciho outperforms KNN, whereas by the PMI expanded datasets, KNN seems to do better.

Table \ref{table:expansion-results} illustrates the effect of the expansion algorithm on the final results. The PMI-based expansion algorithm reduced the mean f-measure of the classifiers. This is because PMI is computed based on available N-gram resources that in the case of Persian, are not pervasive.

\begin{table}[ht]
\centering
\small
\caption{Effect of expansion algorithm on results}
\label{table:expansion-results}
\begin{tabular}{|c|c|}
\hline
Expansion algorithm & F-measure mean \\ \hline
Default & 0.5871 \\ \hline
PMI & 0.5123 \\ \hline
\end{tabular}
\end{table}

The expansion phase has been performed using different combinations of ontology relations. Table \ref{table:expansion-comparison} compares these combinations in the sense of mean f-measure.

\begin{table}[ht]
\centering
\small
\caption{Expansion relations comparison}
\label{table:expansion-comparison}
\begin{tabular}{|c|c|}
\hline
Expansion state & F-measure mean \\ \hline
1 & 0.5184 \\ \hline
2 & 0.5595 \\ \hline
3 & 0.5713 \\ \hline
\end{tabular}
\end{table}

Our results show that the expansion in the third combination performs better than the other two. This means that expanding adjectives with respect to antonym relations leads to a better performance. This observation can be justified with the fact that we are interested in finding words sentiment polarities and adjectives are more likely to express sentiments. This also means that hypernym and hyponym relations between two synsets do not necessarily result in equality between their sentiment polarity. Based on our experiments, these relations show the connection between concepts in the language rather than leading to a better performance in the expansion context.

\subsection{Lexicon generation}
Based on our experiments K-nearest neighbors was selected for lexicon generation. The best training set was Data-5, but in order to include the hand labeled test set in the train data, we concatenated the test data with Data-5 and the result was used for lexicon generation. Table \ref{table:LexiPers} illustrates the total number of synsets in LexiPers and the details of each sentiment orientation.

\begin{table}[ht]
\centering
\small
\caption{LexiPers Details}
\label{table:LexiPers}
\begin{tabular}{|c|c|c|c|}
\hline
\# Synsets & \# Positive & \# Neutral & \# Negative \\ \hline
20432 & 2845 & 14281 & 3306 \\ \hline
\end{tabular}
\end{table}

\section{Conclusion and future work}
\label{sec:conclusion}

In this paper, we introduced a new ontology based method for constructing a sentiment lexicon for Persian. Lexical resources for sentiment analysis in Persian are rather limited. Moreover, most of them uses a translation method for converting the existing lexical resources in other languages like English to Persian. Also, in languages other than Persian, usually works that are based on expanding an initial seed use a previously developed seed for expansion. However, we used a new method for building our own seed, based on finding the strongly connected sub-graphs in the ontology.

As a possible future research, we can apply other learning methods including deep learning to improve the classification accuracy. Moreover, we can assign strength values to the polarity of synsets rather than just assigning a positive or negative label.

\label{sect:bib}
\bibliographystyle{plain}
\bibliography{easychair}

\end{document}